\documentclass[manuscript,nonacm]{acmart}

\AtBeginDocument{%
  }

\usepackage{url}            
\usepackage{booktabs}       
\usepackage{amsfonts}       
\usepackage{nicefrac}       
\usepackage{microtype}      
\usepackage{graphicx}
\usepackage{amsmath, xparse}
\usepackage{multirow}
\usepackage{hhline}
\usepackage{pifont} 
\newcommand{\xmark}{\ding{55}}
\usepackage{flushend}
\usepackage{balance}
\usepackage{subcaption}  

\begin{document}

\title{Quantitative Analysis of Deeply Quantized Tiny Neural Networks Robust to Adversarial Attacks}



%


\author{Idris Zakariyya}
\orcid{0000-0002-7983-1848}
\affiliation{%
  \institution{University of Glasgow}
  \city{Glasgow}
  \country{UK}}
\email{idris.zakariyya@glasgow.ac.uk}

\author{Ferheen Ayaz}
\orcid{0000-0003-3905-675X}
\affiliation{%
  \institution{City, University of London}
  \city{London}
  \country{UK}}
\email{Ferheen.Ayaz@city.ac.uk}
\thanks{This work was done when Ferheen Ayaz was with University of Glasgow.}

\author{Mounia Kharbouche-Harrari}
\orcid{0000-0003-4485-5003}
\affiliation{%
  \institution{STMicroelectronics}
  \country{France}}
\email{mounia.kharbouche-harrari@st.com}

\author{Jeremy Singer}
\orcid{0000-0001-9462-6802}
\affiliation{%
 \institution{University of Glasgow}
 \city{Glasgow}
 \country{UK}}
\email{jeremy.singer@glasgow.ac.uk}

\author{Sye Loong Keoh}
\orcid{0000-0003-3640-5010}
\affiliation{%
 \institution{University of Glasgow}
 \city{Glasgow}
 \country{UK}}
\email{syeloong.keoh@glasgow.ac.uk}

\author{Danilo Pau}
\orcid{0000-0003-1585-2313}
\affiliation{%
  \institution{STMicroelectronics}
  \country{Italy}}
\email{danilo.pau@st.com}

\author{Jos\'e Cano}
\orcid{0000-0002-2243-389X}
\affiliation{%
 \institution{University of Glasgow}
 \city{Glasgow}
 \country{UK}}
\email{josecano.reyes@glasgow.ac.uk}


\renewcommand{\shortauthors}{Zakariyya et al.}



%


\begin{abstract}

Reducing the memory footprint of Machine Learning (ML) models, especially Deep Neural Networks (DNNs), is imperative to facilitate their deployment on resource-constrained edge devices. However, a notable drawback of DNN models lies in their susceptibility to adversarial attacks, wherein minor input perturbations can deceive them. A primary challenge revolves around the development of accurate, resilient, and compact DNN models suitable for deployment on resource-constrained edge devices. 

This paper presents the outcomes of a compact DNN model that exhibits resilience against both black-box and white-box adversarial attacks. This work has achieved this resilience through training with the QKeras quantization-aware training framework. 
The study explores the potential of QKeras and an adversarial robustness technique, Jacobian Regularization (JR), to co-optimize the DNN architecture through per-layer JR methodology. 
As a result, this paper has devised a DNN model employing this co-optimization strategy based on Stochastic Ternary Quantization (STQ). 
Its performance was compared against existing DNN models in the face of various white-box and black-box attacks. The experimental findings revealed that, the proposed DNN model had small footprint and on average, it exhibited better performance than Quanos and DS-CNN MLCommons/TinyML (MLC/T) benchmarks when challenged with white-box and black-box attacks, respectively, on the CIFAR-10 image and Google Speech Commands audio datasets.
    
\end{abstract}

\ccsdesc[500]{Security and privacy}
\ccsdesc[500]{Security and privacy~Systems security}
\ccsdesc[300]{Security and privacy~Distributed systems security}


\keywords{Edge AI, Deep Neural Networks, QKeras, Quantization Aware Training, Jacobian Regularization, Adversarial Attacks.}



\maketitle

\section{Introduction} 
\label{01_intro}


Deep Neural Networks (DNNs) demonstrate remarkable performance in various tasks such as natural language processing (NLP), cybersecurity, computer vision and other heterogeneous applications~\cite{liu2017survey}. 
However, DNN models are resource-intensive, featuring large memory footprints and computational needs~\cite{kim2023c}. Furthermore, the increasing requirements of intelligence at the edge have given rise to new optimization strategies in Machine Learning (ML) which strive to reach high accuracy while shrinking the DNN model architectures at the same time~\cite{gibson_dlas_2024}.
The specific sub-discipline of arttificial intelligence (AI) that generates constrained ML workloads to be deployed on an edge device, such as Microcontroller Units (MCUs) and sensors, is called Edge AI. 


Edge AI offers a promise in terms of executing very low bit depth ML models on resource-constrained MCUs with $\mu$W power and a memory of only a few KiBytes~\cite{TinyML2}. 
Various frameworks such as TensorFlow Lite (TFLite)~\cite{TFlite} for post-training optimization and Larq~\cite{Larq}, Brevitas~\cite{brevitas} and QKeras~\cite{Qkeras} for deep quantization-aware training, are widely used to optimize ML model resources. Most of these frameworks use quantization to optimize the utilized ML models based on data precision. 

QKeras was designed to offer quantization as low as a single bit, and at the same time retain the model accuracy through introducing quantization error in the form of random noise and learning to overcome it during training~\cite{Qkeras2}. It is based on drop-in replacement functions for Keras, 
thus providing the freedom to add a quantizer and choose quantization bit-width separately for activations, biases and weights per layer. This is useful for efficient training of quantized DNN models. 
Among various deep quantization strategies offered by QKeras, there is stochastic quantization~\cite{Sto}, which instead of quantizing all parameters of a DNN model, quantizes a portion of the elements with a stochastic probability inversely proportional to the quantization error, while keeping the other portion unchanged in full-precision (FP). The quantized portion is gradually increased at each iteration until potentially the entire DNN is quantized. This procedure greatly and incrementally compensates the quantization error and thus yields better accuracy for very low-bit-width DNNs. 

However, DNN models, including Edge AI, can be vulnerable to adversarial attacks (causing changes to the input) that are imperceptible to the human eye~\cite{attack1}. These vulnerabilities are critical, as they restrict the deployability of DNN models as an effective solution for real world applications such as autonomous cars, smart cities, intelligent applications and responsive AI~\cite{lin2020adversarial}.
Two widely used classes of adversarial attacks are \textit{white-box} and \textit{black-box} attacks. In white-box, the attacker has full knowledge of the DNN model, its structure and parameters; whereas in the black-box paradigm, the attacker is unaware of the DNN model characteristics. In both scenarios, these attacks aim to cause deliberate mis-classification or to disrupt the model performance. As such, there is an urgent need of balancing the trade-off between reducing the memory footprint of DNN models for tiny edge devices and making them robust against adversarial attacks.

In parallel, exploring the robustness of DNN models is critical to be integrated within Edge AI. In fact, enhancing the security while enabling the model's deployment in resource-constrained MCUs is a key challenge. 
Various defensive techniques for DNNs are present in the literature to provide robustness against adversarial attacks~\cite{silva2020opportunities}. However, such defensive mechanisms may result in an increased model size or accuracy drop for clean sets. Considering Edge AI models, which require extensive learning computation to reach optimal size and accuracy, with possible vulnerability to adversarial attacks, any addition to the model size or drop in accuracy can affect the deployment performance. In view of that, recent studies~\cite {{lin2019defensive,panda2020quanos}} have demonstrated that quantization can reduce computational requirements while granting robustness to a certain level of white-box adversarial attacks. 

Motivated by such an observation, this paper investigates the following question:\\ \textit{Can a per-layer hybrid quantization scheme inherit robustness against white-box and black-box attacks while maintaining the trade-off between clean set accuracy performance and limited-resources requirements of tiny edge devices?}\\ 
This work devised and analyzed some DNN models that are deployable on tiny devices and highly robust to adversarial attacks, trained using the QKeras deep learning framework. The theoretical investigation demonstrated that QKeras utilizes Jacobian Regularization (JR) as an adversarial attack defensive mechanism. 
Based on this, QKeras has been adopted to devise a Stochastic Ternary Quantized (STQ) DNN model with an accuracy performance suitable for deployment in tiny MCUs, and potentially for image and audio processing embedded in the sensor package. Its ability to provide robustness against various adversarial attacks has been investigated. 

The contributions of this paper include the following:

\begin{itemize}
  \item Enhancement of the STQ-based DNN model with low bit-width parameters proposed in~\cite{ayaz_ijcnn}, so that can be deployed on MCUs with minimal memory footprint requirements and an improved accuracy performance on clean sets compared with other QKeras quantized models (Section~\ref{appr}).
  
  \item A comprehensive comparison of the resilience of the STQ-based model to industry standard MLCommons/TinyML (MLC/T) benchmarks, the existing quantized model Quanos, RND-AT and Bit-Plane solutions, for both image and audio inputs; while investigating its performance with $K$-fold cross validation (Section~\ref{eval}).

  \item An analysis against the state of art model MobileNetV2 modified by adopting the STQ design methodology, for cross comparison.


 
  
\end{itemize}

The rest of the paper is organized as follows: Section~\ref{backrw} discusses the background and related work. Section~\ref{appr} presents the analysis of the defensive mechanism of QKeras against adversarial attacks, while Section~\ref{eval} describes the implementation details and results. Section~\ref{conc} concludes the paper and gives some ideas for future work.

\section{Background and Related Work} 
\label{backrw}

This section reviews the background information related to adversarial attacks and some solutions and mitigations that have been developed to improve the robustness of DNN models against these threats. Additionally, it discusses the related works of JR and robustness of quantized models in existing literature.


\subsection{Adversarial Attacks Against Machine Learning Models}

Security is very important for sensitive and embedded applications like intelligent transportation systems, home, consumer, automotive, healthcare and financial systems~\cite{hamdi2023enhancing}. On October 21, 2016, three consecutive distributed denial-of-service attacks were launched against the Domain Name System (DNS) provider Dyn~\cite{alieyan2016overview}. The attack caused major Internet platforms and services to be unavailable in Europe and North America. The activities were believed to have been executed by means of many internet-connected devices—such as printers, IP cameras, residential gateways, home appliances and baby monitors—that had been infected with the Mirai malware~\cite{antonakakis2017understanding}. That experience suggested that internet of things (IoT) devices which utilize image and voice recognition AI-based models~\cite{sikder2021survey} were at high risk of attacks which compromised their integrity, confidentiality and availability in real world applications. 
In particular, adversarial attacks involve adding a small perturbation to the input to maximize the loss function of a model under a constrained norm~\cite{su2019one}. 
Equation (\ref{eqadp}) expresses such a procedure of introducing a perturbation into an input data, where: $\mathcal{L}(\theta, x', y)$ is the loss function, $\theta$ denotes the model parameters, $x'$ is the perturbed input, $y$ is the model output, $\delta$ denotes the perturbation and $p$ is the perturbation norm~\cite{jiang2021project}. This work considered two types of attacks, white-box and black-box attacks.

A common adversarial attack technique is the Fast Gradient Sign Method (FGSM)~\cite{goodfellow2014explaining}. This white-box attack uses a single-step iteration to estimate the gradient of the model training loss function, based on the inputs. An FGSM attack procedure is expressed in (\ref{fgsm}), where $\Delta$ represents the gradient and $\epsilon$ denotes a small constant value that restricts the perturbation. 
A variation of FGSM is the Projected Gradient Descent (PGD)~\cite{madry2017towards}. This is a more computationally expensive multi-step threat model which runs several iterations to find an adversarial input with the lowest possible $\delta$, as expressed in (\ref{pgd}); where $i$ is the iteration index, $\alpha$ denotes the gradient step size and $S$ represents the perturbation set. 
Moreover, Carlini and Wagner (C\&W) is another white-box attack introduced to thwart state-of the art defenses, with a better success rate than FGSM and PGD attack methods~\cite{carlini2017towards}. This new algorithm has been proven to be successful on both distilled and un-distilled neural networks.
\begin{equation}
\label{eqadp}
\smash{\displaystyle\max_{|| \delta||_p}}\, \mathcal{L}(\theta, x', y)   
\end{equation} 
\begin{equation}
\label{fgsm}
    x' = x + \epsilon \cdot sign(\Delta_x \mathcal{L}(\theta, x, y))
\end{equation}
\noindent
\begin{equation}
\label{pgd}
    x_{i+1}' = \prod_{x+S} (x + \alpha \cdot sign(\Delta_x \mathcal{L}(\theta, x, y)))
\end{equation}

\begin{figure*}[t]
	\centering
	\includegraphics[scale=0.45]{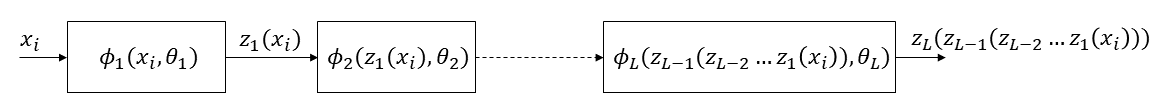}
	\caption{Transformation of input $x_i$ into output $z_L$ by DNN.}
	\label{fig:transformation}
\end{figure*}

An efficient black-box attack is the Square attack~\cite{square}, which is based on a random search optimization technique with multiple iterations. In each iteration, it changes a small fraction of the input shaped into squares at random positions. Similar to gradient-based optimizations, it also relies on step-size reduction, where the size refers to the dimensions of the square~\cite{square}. 
Another form of black-box attack is the Boundary attack~\cite{boundary}, where the queries are used to estimate the decision boundaries of the output classes. A Boundary attack starts with a clean image and the gradient estimation is performed with queries, moving along the estimated direction in each iteration and projecting a new perturbation until the model decision is changed~\cite{boundary}. A stronger black-box attack is the Zeroth Order Optimization (Zoo) attack that directly estimates the gradients of the targeted DNN models whilst generating adversarial inputs~\cite{chen2017zoo}.

Overcoming the white-box and black-box attack methods requires a suitable defensive mechanism. Therefore, it is important to enhance the model robustness against different attack variations to ease their deployment.





\subsection{Robustness against Adversarial Attacks}

Various defense methods have been proposed to increase the robustness of DNNs against adversarial attacks~\cite{xu2020adversarial}. Some strategies aim at detecting adversarial inputs~\cite{detect} or performing transformations to remove perturbations through an additional network or module~\cite{transform}-\cite{Bernhard_LuringAE_IJCNN21}. 
Adversarial training introduces adversarial inputs during the model's learning so that it learns not to misinterpret them~\cite{AT}. However, these approaches do not guarantee robustness against attacks which are not introduced to the model during training, as they are not learned by the model. 
Other methods such as~\cite{picot2022adversarial} decrease the model's sensitivity to small perturbations by adding a regularization term in the loss function. Equation (\ref{joinl}) expresses this joint loss function, where $\mathcal{L}_{reg}$ denotes the regularization term and $\lambda$ is a hyper-parameter used to allow the adjustment between the regularization and the actual loss.  
\begin{equation}
\label{joinl}
\mathcal{L}_{joint} = \mathcal{L}(\theta, x, y) + \mathcal \lambda{L}_{reg},  
\end{equation}


\subsection{Jacobian Regularization}

JR is a technique that provides adversarial robustness, where the input regularization gradient normalizes the gradient of the cross-entropy loss, as expressed in Equation (\ref{L_reg}): 
\begin{equation}
\label{L_reg}
\mathcal{L}_{reg} = ||J(\mathbf{x}_i)||_F^2
\end{equation}
where $||J(\mathbf{x}_i)||_F^2$ is the Frobenius norm of the model’s Jacobian matrix evaluated on the input data~\cite{Jacobian1}. Per-layer JR is proposed in~\cite{Jacobian2} to reduce the computational complexity of the original JR. 

To demonstrate the robustness of per-layer JR, lets consider a $D$-dimensional network input $X$ consisting of $N$ training samples. As shown in Figure~\ref{fig:transformation}, the network contains $l=1,2,...,L$ layers. 
$z_l$ denotes the output at layer $l$ and $z_l^k$ is the output of the $k^{th}$ neuron of the $l^{th}$ layer. Consider the softmax operation at the output of the network; the predicted final output for computing the top-1 accuracy for an input $x_i$ is $f(x_i)=argmax \{ z_L^1, z_L^2,..,z_L^K \}$, where $K$ is the dimension of the output vector. 
The Jacobian matrix of a DNN is computed at $L^{th}$ layer, i.e. $J_L(x_i) = \nabla_x z_L (x_i)$, and is defined as:
\begin{equation}
	J_L (x_i) = 
	\begin{bmatrix}
		\frac{ \partial z_L^1 x_1 }{ \partial x_1} & \frac{ \partial z_L^1 x_1 }{ \partial x_2} & ... & \frac{ \partial z_L^1 x_1 }{ \partial x_D} \\
		\frac{ \partial z_L^2 x_2 }{ \partial x_1} & \frac{ \partial z_L^2 x_2 }{ \partial x_2} & ... & \frac{ \partial z_L^2 x_2 }{ \partial x_D} \\
		. & . & . & . \\
		. & . & \;\; . & . \\
		\frac{ \partial z_L^K x_K }{ \partial x_1} & \frac{ \partial z_L^K x_K }{ \partial x_2} & ... & \frac{ \partial z_L^K x_K }{ \partial x_D} \\
			
	\end{bmatrix} \epsilon \mathbb{R} ^{K \times D}.
\end{equation}

In~\cite{Jacobian1}, the JR term for an input $x_i$ is defined as
\begin{equation}
\small
	||J(x_i)||_F^2 = \sum_{d=1}^D \sum_{k=1}^K \Big( \frac {\partial} {\partial x_d} z_L^k (x_i) \Big)^2 = \sum_{k=1}^K ||\nabla_x z_L^k (x_i) ||^2_2,
\label{Jreg}
\end{equation}

The standard loss of the training is added with the regularization term in (\ref{Jreg}) to improve the robustness of the DNN. It is proposed as a post-training, in which the network is re-trained for fewer iterations with the new loss function~\cite{Jacobian1}. As $i \epsilon \{1,2,...,N\}$, JR requires the computation of $N$ gradients, whereas in per-layer JR, the Jacobian matrix is computed on only one random $i$ at each layer~\cite{Jacobian2}. 
The basic idea is to reduce the Frobenius norm of the Jacobian matrix, which results in the expansion of the classification margin, i.e. the distance between an input and the decision boundary induced by a network classifier.


\subsection{Adversarial Robustness of Quantized Models}

Model quantization techniques are widely used in various fields~\cite{gholami2021survey}. 
Pau et al.~\cite{online2} uses it to reduce memory footprint and computational costs for image detection, audio classification and thwarting cyber attacks in IoT networks. 
However, the limitation of a quantized model is the shift of the FP model classification boundary, which may influence how vulnerable the model is to adversarial perturbations~\cite{song2020improving}. 
As such, Song et al.~\cite{song2020improving} investigated the use of a boundary-based retraining method to reduce adversarial and quantization losses with the usage of non-linear mapping as a defensive mechanism against white-box adversarial attacks. 

Other previous studies explored the impact of perturbations on different models. Table~\ref{mlcompar} compares some existing works that investigated the robustness of quantized DNN models. 
Panda.~\cite{panda2020quanos} proposed a hybrid quantization framework based on adversarial noise sensitivity. It determines the ideal bitwidth for DNNs and its relevance to adversarial robustness. It has been evaluated on CIFAR-10 datasets to assess the robustness of FGSM and PGD across various perturbation strengths. Liu et al.~\cite{liu2022defending} proposed a defensive mechanism for DNNs that combined a target DNN classifier with bit-plane classifiers, extracting bit-slices from input images. This technique was assessed against white-box and black-box attacks on CIFAR-10 dataset, including C\&WL2, Zoo and Boundary attacks methods. 
A lightweight defense strategy aimed at mitigating query-based attacks was proposed in~\cite{qin2021random}. For the evaluation, it was considered CIFAR-10 against query-based black-box attacks, including Square under different perturbation norms. 
These studies purely focused on the resilience of quantized DNN models without examining their deployment feasibility on resource-constrained devices, such as MCUs or sensors, and some of them considered only white-box attacks, which are less interesting with respect to black-box ones as most of the industry deployments do not make public the model's topology embedded in its products. 

Therefore, this paper investigates the effectiveness of the STQ methodology developed, in terms of robustness against various attack strengths, for widely used benchmarks~\cite{panda2020quanos,liu2022defending, qin2021random} and demonstrates its robustness capability, with a small memory footprint. 
In addition, STQ was evaluated using defensive distillation, a mechanism that mitigates white-box and black-box attacks~\cite{papernot2017extending}, by smoothing the learned model, using a temperature $T$ parameter in the computation of the softmax activation function. The chosen temperatures for the defensive evaluation are $1$ (no distillation applied) and $50$.

\begin{table}[!tp]
\caption{Comparison of adversarially quantized DNN models.}
\centering
\fontsize{6.5}{9}\selectfont
\label{mlcompar}
\begin{tabular}{|cc|cccc|c|}
\hline
\multicolumn{2}{|c|}{\textbf{Works}} &
  \multicolumn{1}{c|}{\cite{galloway2017attacking}} &
  \multicolumn{1}{c|}{\cite{bernhard2019impact}} &
  \multicolumn{1}{c|}{\cite{kim2020robust}} &
  \cite{gorsline2021adversarial} &
  Ours \\ \hline
\multicolumn{1}{|c|}{\multirow{3}{*}{\begin{tabular}[c]{@{}c@{}}\textbf{Compress.} \\ \textbf{technique}\end{tabular}}} &
  Method &
  \multicolumn{4}{c|}{N bit-width quantization} &
  SQ* \\ \cline{2-7} 
\multicolumn{1}{|c|}{} &
  on weights &
  \multicolumn{1}{c|}{\checkmark} &
  \multicolumn{1}{c|}{\checkmark} &
  \multicolumn{1}{c|}{\checkmark} &
  \checkmark &
  \checkmark \\ \cline{2-7} 
\multicolumn{1}{|c|}{} &
  on activat. &
  \multicolumn{1}{c|}{\checkmark} &
  \multicolumn{1}{c|}{\checkmark} &
  \multicolumn{1}{c|}{\checkmark} &
  \xmark &
  \checkmark \\ \hline
\multicolumn{1}{|c|}{\multirow{2}{*}{\textbf{Inputs}}} &
  Images &
  \multicolumn{1}{c|}{\checkmark} &
  \multicolumn{1}{c|}{\checkmark} &
  \multicolumn{1}{c|}{\checkmark} &
  \checkmark &
  \checkmark \\ \cline{2-7} 
\multicolumn{1}{|c|}{} &
  Audio &
  \multicolumn{1}{c|}{\xmark} &
  \multicolumn{1}{c|}{\xmark} &
  \multicolumn{1}{c|}{\xmark} &
  \xmark &
  \checkmark \\ \hline
\multicolumn{2}{|c|}{\textbf{Datasets}} &
  \multicolumn{1}{c|}{\begin{tabular}[c]{@{}c@{}}MNIST\\ C-10*\end{tabular}} &
  \multicolumn{1}{c|}{\begin{tabular}[c]{@{}c@{}}MNIST\\ SVHN\end{tabular}} &
  \multicolumn{1}{c|}{\begin{tabular}[c]{@{}c@{}}MNIST\\ C-10*\\ TinyI*\end{tabular}} &
  \begin{tabular}[c]{@{}c@{}}MNIST\\ Spiral\end{tabular} &
  \begin{tabular}[c]{@{}c@{}}C-10*\\ SVHN\\ GSC\end{tabular} \\ \hline
\multicolumn{1}{|c|}{\multirow{6}{*}{\textbf{Attacks}}} &
  \begin{tabular}[c]{@{}c@{}}White\\ box\end{tabular} &
  \multicolumn{1}{c|}{\begin{tabular}[c]{@{}c@{}}FGSM\\ PGD\\ C\&W\end{tabular}} &
  \multicolumn{1}{c|}{\begin{tabular}[c]{@{}c@{}}FGSM\\ BIM\\ C\&W\end{tabular}} &
  \multicolumn{1}{c|}{\begin{tabular}[c]{@{}c@{}} \xmark \end{tabular}} &
  FGSM &
  \begin{tabular}[c]{@{}c@{}}FGSM\\ PGD\\ C\&W\end{tabular} \\ \cline{2-7} 
\multicolumn{1}{|c|}{} &
  \begin{tabular}[c]{@{}c@{}}Black\\ box\end{tabular} &
  \multicolumn{1}{c|}{\xmark} &
  \multicolumn{1}{c|}{\begin{tabular}[c]{@{}c@{}}SPSA \\ ZOO \end{tabular}} &
  \multicolumn{1}{c|}{\xmark} &
  \xmark &
  \begin{tabular}[c]{@{}c@{}}Square\\ Boundary\\ ZOO\end{tabular} \\ \cline{2-7}
\multicolumn{1}{|c|}{} &
  Random &
  \multicolumn{1}{c|}{\xmark} &
  \multicolumn{1}{c|}{\xmark} &
  \multicolumn{1}{c|}{\begin{tabular}[c]{@{}c@{}}Gaus.\\ noise\end{tabular}} &
  \xmark & \xmark
   \\ \hline
  
\multicolumn{2}{|c|}{\textbf{Memory footprint}} &
  \multicolumn{1}{c|}{\xmark} & \multicolumn{1}{c|}{\xmark} & \multicolumn{1}{c|}{\xmark} & \xmark & 410 KB \\ \hline
\multicolumn{7}{l}{*C-10 denotes CIFAR-10 dataset, *TinyI: TinyImageNet, *SQ: Stochastic Quantization}
\end{tabular}
\end{table}

\section{Proposed Deep Neural Network model} 
\label{appr}


This section introduces the investigation of the newly devised and robust DNN model, with minimized memory requirements, and suitable for deployment on constrained devices. This model is based on the STQ approach and was conceived to be deployable on tiny MCUs.


\subsection{Adversarial Robustness in QKeras}

\begin{figure}[tp]
  \centering
  \includegraphics[width = 0.6\columnwidth]{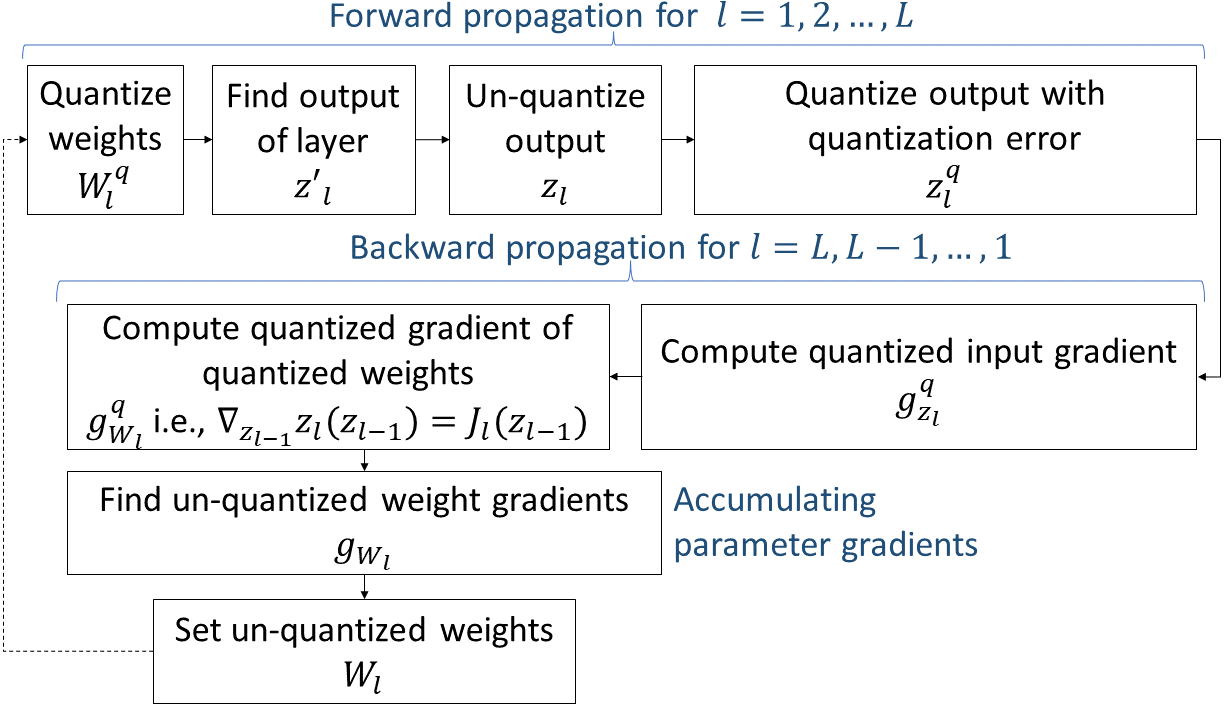}
   \caption{QKeras quantization aware training flow.}
 	\label{qkerasalgo}
\end{figure}

QKeras~\cite{Qkeras} is a DNN framework featuring activations and parameters quantization through drop-in function replacements on Keras~\cite{keras},
which provides a productive, easy-to-learn methodology to build and train quantized neural networks, either fractional or integer spanning from $1$ to $32$ bits. 
QKeras performs quantization-aware training~\cite{qkeras3}. The background algorithm to train neural networks with $q$ bit-width weights, activations and gradient parameters is conceptualized in~\cite{dorefa}. 
In particular, the network training in QKeras includes a backward propagation where parameters are stochastically quantized into low bit-width numbers. 
Figure~\ref{qkerasalgo} shows the process flow of training each layer in QKeras. Each training iteration involves a forward propagation step to quantize the weights, find the corresponding output and add the quantization error into the output of each layer. 
Finally, unquantized weights and gradients are updated for the next iteration. The training process of deeply quantized networks, through QKeras, make them robust to adversarial attacks due to the two following reasons: i) a noise function is introduced during quantization of gradients to overcome quantization error during training, which makes a DNN robust to noise and perturbation effect; ii) QKeras involves JR features as explained below.


\textbf{Theorem:} \textit{QKeras introduces per-layer JR and therefore increases the robustness of DNNs against adversarial attacks.}

\textbf{Proof:} Considering the per-layer structure of DNNs, as shown in Figure~\ref{fig:transformation}, the output $z_L$ at the last layer $L$ is
\begin{equation}
	z_L = \phi_L(\phi_{L-1}(...\phi_1(x_i, \theta_1),...\theta_{L-1}), \theta_L)
\end{equation}
where $\phi_l(., \theta_l)$ represents the function of the $l^{th}$ layer, $\theta_l$ denotes the model parameters at layer $l$, and $z_0 = x_i$~\cite{Jacobian2}. The Jacobian matrix of the $l^{th}$ layer is defined as 
\begin{equation}
J_l(z_{l-1}) = \frac{dz_l}{dz_{l-1}},
\label{back}
\end{equation}
which is back-propagated during QKeras training (Step 10-16 of Algorithm 1 in~\cite{dorefa}). The derivation expressed in (\ref{back}) is the Jacobian matrix~\cite{Jacobian2} of the $l^{th}$ block layers. Thus, QKeras with such patterns incorporates per-layer JR during training.


To prove that per-layer JR enhances adversarial robustness, consider a clean input $x_c$ and an adversarial input $x_p$, both close to an input $x_i$ and all belonging to the same class $k$. Since $f(x_i) = f(x_c) \neq f(x_p)$, then the $\ell_2$ distance metric of the input and output of the network, as defined in~\cite{Jacobian1}, is:
\begin{equation}
	\frac{||x_p - x_i||_2}{||x_c - x_i ||_2}  \approx 1,
\label{p1}
\end{equation}

\begin{equation}
	\frac{||z_L (x_p) - z_L (x_i)||_2}{||z_L (x_c) - z_L (x_i) ||_2}  > 1,
\label{p2}
\end{equation}
Combining (\ref{p1}) and (\ref{p2}) and using the Mean Value Theorem~\cite{shishkina2022mean}, it is justified that a lower Frobenius norm makes a network less sensitive to perturbations, i.e.,
\begin{equation}
\frac{||z_L (x_p) - z_L (x_i)||_2^2}{||x_p - x_i ||_2^2}  \leq ||J(x')||_F^2,
\end{equation}
where $x' \epsilon [x_i, x_p]$. Similar to (\ref{p2}), for each layer $l$, we have
\begin{equation}
	\frac{||z_l (x_p) - z_l (x_i)||_2}{||z_l (x_c) - z_l (x_i) ||_2}  > 1.
	\label{p3}
\end{equation}
The misclassification error is propagated at each layer, thus
\begin{equation}
	\frac{||z_l (z_{l-1}(x_p)) - z_l (z_{l-1}(x_i))||_2}{||z_l (z_{l-1}(x_c)) - z_l (z_{l-1}(x_i)) ||_2}  > 1.
	\label{p4}
\end{equation}

This error is back-propagated and then adjusted at each layer. Therefore, the learning optimization process increases the robustness of a DNN model trained by QKeras, since it discriminates the error due to the clean versus the perturbed input. 
Consequently, it can be concluded that QKeras deep quantization-aware process introduces per-layer JR with respect to the previous layer’s input, that is back-propagated, as shown in Figure~\ref{qkerasalgo} and Step 11 to 15 of Algorithm 1 in~\cite{dorefa}. 
Although back-propagation is intended to quantize parameters, it ultimately results in more robust networks. Even though this learning  process increases the training computation time, its advantages are two-fold, i.e. deep quantization and robustness to adversarial perturbations.


\subsection{Stochastic Ternary Quantized Architecture}

\begin{figure*} [!tp]
  \centering
  \includegraphics[width=0.95\textwidth]{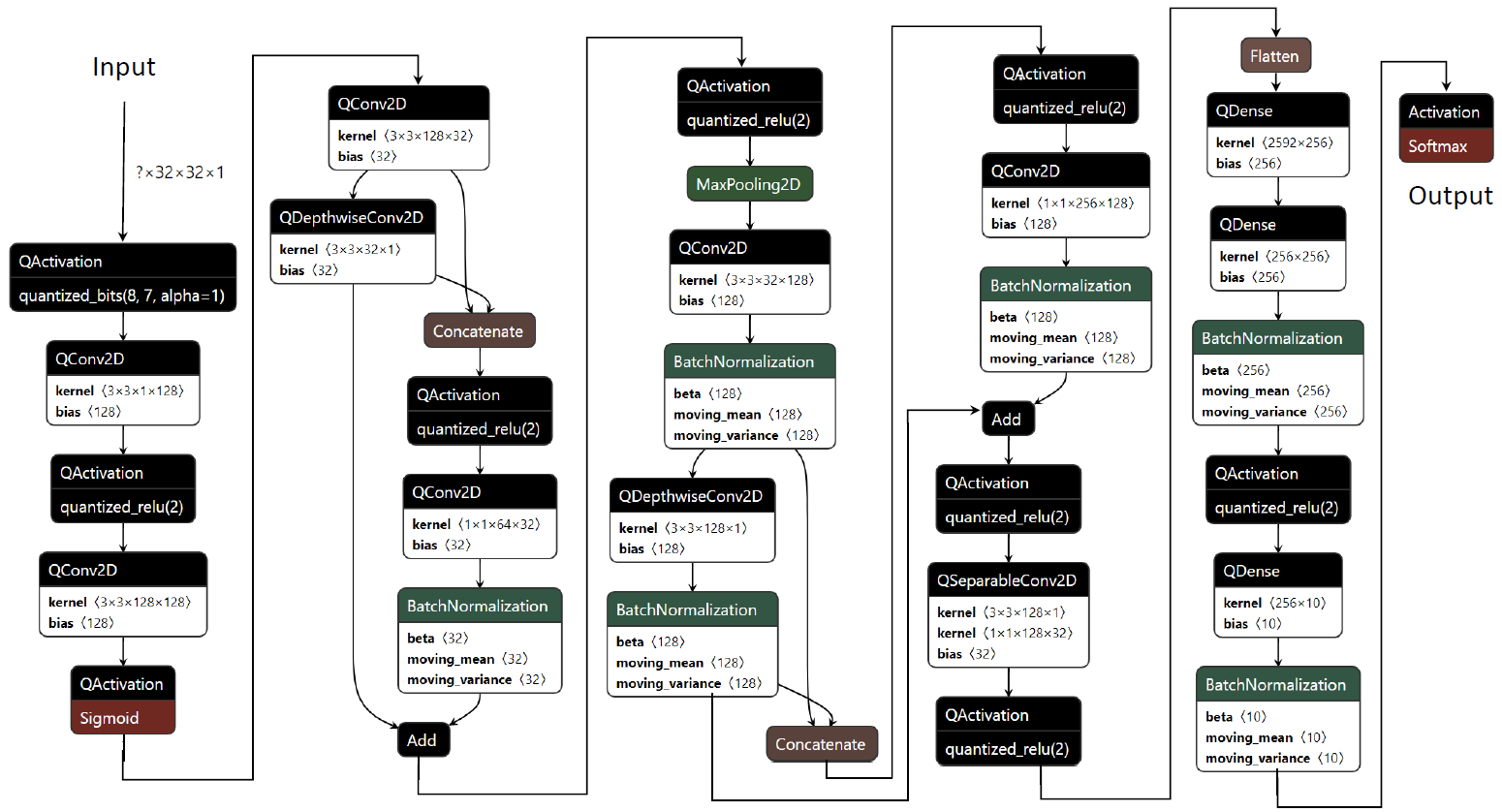}
   \caption{Proposed architecture of the STQ-based DNN model designed and trained with QKeras.}
   \label{fig3}
\end{figure*}

As the QKeras~\cite{qkeras3} API serves as an extension of Keras, initially a 32-bit FP DNN model was built with Keras~\cite{keras}. The FP model consists of six convolutional layers including depth-wise or separable configurations, and three fully connected layers. Former layers are used for capturing channel-wise correlations and can provide more features with less parameters, particularly with image inputs~\cite{guo2019depthwise}. 
A multi-branch topology was used with residual connections to refine the feature maps. For better accuracy, batch normalization~\cite{wang2021enabling}, ReLU~\cite{eckle2019comparison} and sigmoid~\cite{papernot2021tempered} activation functions are considered in the hidden layers, while softmax is used in the output layer. 
The FP model contains $998,824$ parameters, of which $997,460$ are trainable and $1,364$ are non-trainable. This FP DNN model is not integrated with JR by default, and is therefore vulnerable to adversarial attacks. Moreover, the model cannot be deployed on tiny MCUs due to its large size (4MB). Since this paper targets resource-constrained edge devices, the selected MCU was the STM32H735GDK nucleo IoT board, which featured embedded SRAM memory of 564KiB and FLASH of 1MB. The STQ method has been applied to this model by using QKeras.

Figure~\ref{fig3} shows the architecture of the STQ model as rendered by Netron~\cite{netron} to inspect the DNN graph along with its hyperparameters. As the figure shows, the convolutional, depth-wise, separable layers and activation are appended with \textit{Q}, which indicates the quantization version of the FP Keras~\cite{keras} layers, while the \textit{quantized\_relu} represents the quantized activation function version of FP Keras's ReLU. The proposed STQ model uses the heterogeneous quantization features of QKeras, which support independent quantization of each layer in the DNN~\cite{Qkeras2}. This is useful in reducing the model's memory footprint and complexity, while increasing accuracy.




\section{Evaluation} 
\label{eval}

This section describes the evaluation procedure of the proposed STQ model, using image and audio datasets for clean and adversarial samples. Moreover, it compares the top-1 accuracy of the proposed STQ model with other benchmarks for three black-box (Square, Boundary and ZOO attacks) and three white-box (FGSM, PGD and C\&W) adversarial attacks.


\subsection{Experimental Setup}

\subsubsection{Datasets and Pre-processing}

To evaluate the effectiveness of the devised STQ model, the following audio and image benchmark datasets were considered:

\begin{itemize}
  \item \textbf{CIFAR-10} consists of $60,000$ images belonging to $10$ different classes. This dataset is divided into $50,000$ training images and $10,000$ test images~\cite{cifar10}.
  
  \item \textbf{Street View House Numbers (SVHN)} is a real-world image dataset capturing house numbers in Google Street View images. It consists of $10$ classes, one for each digit. There are $73,257$ and $26,032$ digits for training and testing respectively~\cite{SVHN}.
  
  \item \textbf{Google Speech Commands (GSC)} consists of $65,000$ one-second long utterances of $30$ short words by thousands of different people. Its $12$ classes comprise words of `yes', `no' and digits from `zero' to `nine' that were used in the experiment. The number of training and test samples are $31,257$ and $15,636$, respectively~\cite{GSC}.
\end{itemize} 

Note that the input image samples were normalized between 0 and 1 pixels and then converted into gray-scale images with pixel values ranging from [-128 to +127], before presenting them to the DNN input. This effectively reduced the computational cost of processing the image samples while avoiding the use of colors which are known to be deceptive, and because it is well known that the most signal's energy part is embodied into the luminance component. Therefore, color processing was out of scope in this work and may be considered as future extensions for problems where the color components are essential.


\subsubsection{Model Training Procedure}

Table~\ref{tableparam} lists the training hyperparameters used to evaluate the tested quantized models. A cosine annealing Learning Rate (LR) function~\cite{loshchilov2016sgdr} was used with the Adamax optimizer for faster convergence, except for the FP, ternary and STQ for which the Root Mean Square (RMS) optimizer is considered for faster convergence~\cite{postalciouglu2020performance}. These hyperparameters were selected to both fine-tune each model and reduce its computational complexity, while providing better or maintaining state-of-the-art performance across all the tested DNN models.

\begin{table}[!ht]
\caption{Training parameters for QKeras DNN.}
  \label{tableparam}
  \centering
  \begin{tabular}{|c|c|}
  \hline
  \textbf{Parameter} & \textbf{Value} \\ \hline
  {Epochs} & {1000} \\ \hline
  {Batch Size} & {64} \\ \hline 
  {Learning Rate} &  [1 $\times$ $10^{-6}$, 1 $\times$ $10^{-3}$] \\ \hline
  {LR Scheduler} & {Cosine Annealing} \\ \hline
  {Loss} & {Categorical Cross-entropy} \\ \hline
  {Optimizer} & {Adamax or RMS} \\ \hline
 
  \end{tabular}
\end{table}

Furthermore, a comparative analysis was carried out involving industry-standard MLC/T ResNetv1 and DS-CNN models performance against the proposed FP and STQ models. This approach allowed to investigate the models robustness through various attacks. The inclusion of these well-known benchmarks offered valuable insights into the performance potential of the proposed model and methodology across heterogeneous datasets. To ensure a fair comparison, an equal number of training epochs were utilized, mirroring the training procedure applied to STQ. 
Codes for empirical evaluations of these different models through different attacks are available on github~\footnote{\url{https://github.com/izakariyya/STQ/tree/main}}.



\subsubsection{Adversarial Attacks Procedure}

The proposed STQ model was evaluated against several adversarial attacks to demonstrate its resilience and robustness. Three white-box attacks, namely FGSM~\cite{goodfellow2015explaining}, PGD~\cite{madry2019deep} and C\&W (L2)~\cite{carlini2017towards}; and three black-box attacks, namely Square~\cite{square}, Boundary~\cite{boundary} and Zero Order Optimization (Zoo)~\cite{chen2017zoo}, were considered. 
The perturbed data samples for all attacks were generated with the Adversarial Robustness Toolbox (ART)~\cite{nicolae2018adversarial} against the tested datasets. For FGSM samples, an $\epsilon$ value of $0.3$ was used while for PGD an $\epsilon$ value of $32/255$, with a maximum iteration value of $7$ was used. 
For the Square attack, an $\epsilon$ value of $0.3$ with a maximum number of iterations of $10$ was used, while for the Boundary attack, the non-targeted technique was adopted with an $\epsilon$ value of $0.01$ and a maximum number of iterations of $5000$, for better attack strength. The number of samples used to create the perturbations were $10,000$, $26,032$ and $15,636$ for CIFAR-10, SVHN and GSC datasets, respectively. 


To further examine the strength of the STQ model against FGSM, PGD~\cite{huang2020bridging} and Square~\cite{square} attacks, they were crafted under different attack strengths using the $10,000$ test samples of CIFAR-10. 
For the FGSM attacks, adversarial examples were created for varying values of $\epsilon$ in the range ${0.05, 0.1, 0.15, 0.2, 0.25, 0.3}$, as described in~\cite{panda2020quanos}. Regarding the PGD attack, an iteration $t = 7$ was used along with a step size $\alpha = 2/255$ and an $\epsilon$ ranging in: $\{8/255, 16/255, 32/255\}$ as implemented in~\cite{panda2020quanos}. This is to compare the proposed STQ with QUANOS model~\cite{panda2020quanos}. The QUANOS model utilizes random hybrid quantization. For example, given a 16-bit precision DNN, QUANOS applies hybrid quantization based on the sensitivity of each layer. The employed model for this task was VGG-19.

For the Square~\cite{square} attacks, two variations were utilized. The first variation employed the \emph{infinity} norm with an $\epsilon$ value of $0.05$ and a maximum of $10,000$ iterations. The second variation used is based on the $\ell_2$ norm with a maximum of $10,000$ iterations, as specified in~\cite{qin2021random}. The model utilized in~\cite{qin2021random} is a lightweight VGG-16, with randomly added Gaussian noise for defense against query-based black box attacks

STQ was further compared with Bit-Plane~\cite{liu2022defending} tested with adversarial samples generated by $C\&W-L_{2}$, Boundary and ZOO attacks. In each case, $2000$ random samples of CIFAR-10 were utilized to create the perturbations. 
The Bit-Plane classifier, as described in~\cite{liu2022defending}, operates by employing bit-planes of input images for convolution, focusing particularly on the low-order bits where imperceptible perturbations from attacks tend to occur. Initially, RGB images are inputted into the classifier, and a bit slicing of $24$ bits is performed. For details on the model architecture used in this approach, refer to~\cite{qiu2020mitigating}.


\subsubsection{$K$-fold Cross Validation}

To assess the mean accuracy and standard deviation of both clean and adversarial data samples across each model under consideration, a $K$-fold cross-validation approach was employed. This methodology involves dividing the complete dataset into $K$ equal subsets, where $K-1$ of these subsets are allocated for training purposes, and the remaining subset is utilized for validation, as detailed in~\cite{berrar2019cross}.

For each fold in the cross-validation process, the model is trained on the training set and subsequently tested on the validation set to determine the average performance. To limit the computational complexity, a value of $K = 5$ was selected, aligning with the conventional practice, as it is commonly used for assessing predictive performance, as discussed in~\cite{jung2018multiple} and~\cite{wong2019reliable}. Additionally, the scikit-learn machine learning Python API, as documented in~\cite{sklearn}, was employed for the implementation of this cross-validation technique.


\subsection{Quantization schemes}

Table~\ref{table:QKeraseval} shows the performance (top-1 test accuracy) comparison between different quantized models using clean inputs from each dataset. In addition, it includes the FLASH MCU memory of each quantized model computed based on the weight profile obtained using the \texttt{qstats} QKeras library. 
It is observed that the STQ-based model provides the highest accuracy compared to the FP, Stochastic Binary (S-Binary) and other quantized models. Moreover, this model requires only $410$KB being lighter than the FP, 8-bits and 4-bits models thus making its footprint deployable in the MCU's embedded memory, such as STM32. These results motivate further investigations into the robustness of the proposed STQ model against adversarial attacks, since the FP model does not have any integrated defensive mechanism. 

\begin{table}[!t]
\caption{Performance (top-1 accuracy) comparison on CIFAR-10, SVHN and GSC clean datasets.}
\begin{center}
\label{table:QKeraseval}
\begin{tabular}{|c|c|c|c|c|c|c|}
\hline
\textbf{\multirow{2}{*}{Model}} & \textbf{Flash} & \textbf{CIFAR-10}  & \textbf{SVHN} & \textbf{GSC} \\
& \textbf{(KB)} & \textbf{Acc. (\%)}  & \textbf{Acc. (\%)}  & \textbf{Acc. (\%)}          \\ \hline
{FP} & {4496} & {75.20} &{74.27} &{90.55}   \\ \hline
{8-bit}   & {1124} & {33.82} &{46.29} &{37.87}   \\ \hline
{4-bit}  & {527} & {50.18} &{63.04} &{76.95}   \\ \hline
{Ternary}  & {410} & {62.13} &{93.49} &{89.9} \\ \hline
{STQ-T50} & {410} & \textbf{80.57} &\textbf{94.33} &{88.85} \\ \hline
{STQ-T1} & {410} &{75.58} &{94.29} &\textbf{90.55} \\ \hline
{2-bit} & {281} & {62.85} &{92.67} &{88.03}   \\ \hline
{Binary}   & {140} & {37.30} &{59.27} &{61.88} \\ \hline
{S-Binary} & {140} & {51.19} &{65.13} &{76.61}\\
\hline
\end{tabular}
\end{center}
\end{table}

\subsection{Robustness evaluation against adversarial attacks}

\begin{figure*} 
\centering
  \begin{subfigure}[b]{0.70\linewidth}
    \centering
    \includegraphics[width=0.99\linewidth]{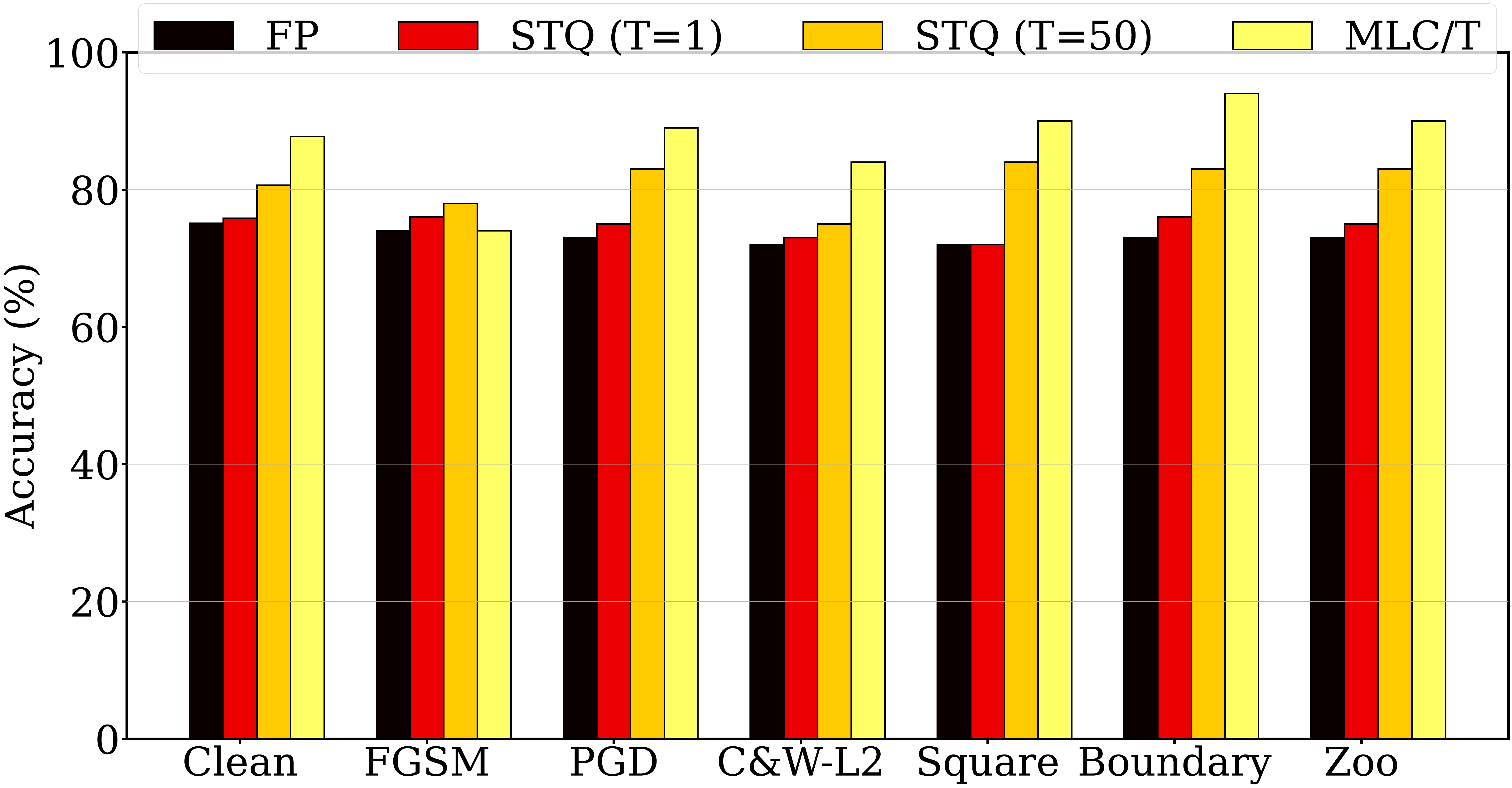}
    \caption{CIFAR-10}
    \label{fig:cif_kf}
  \end{subfigure}
  \begin{subfigure}[b]{0.70\linewidth}
    \centering
    \includegraphics[width=0.99\linewidth]{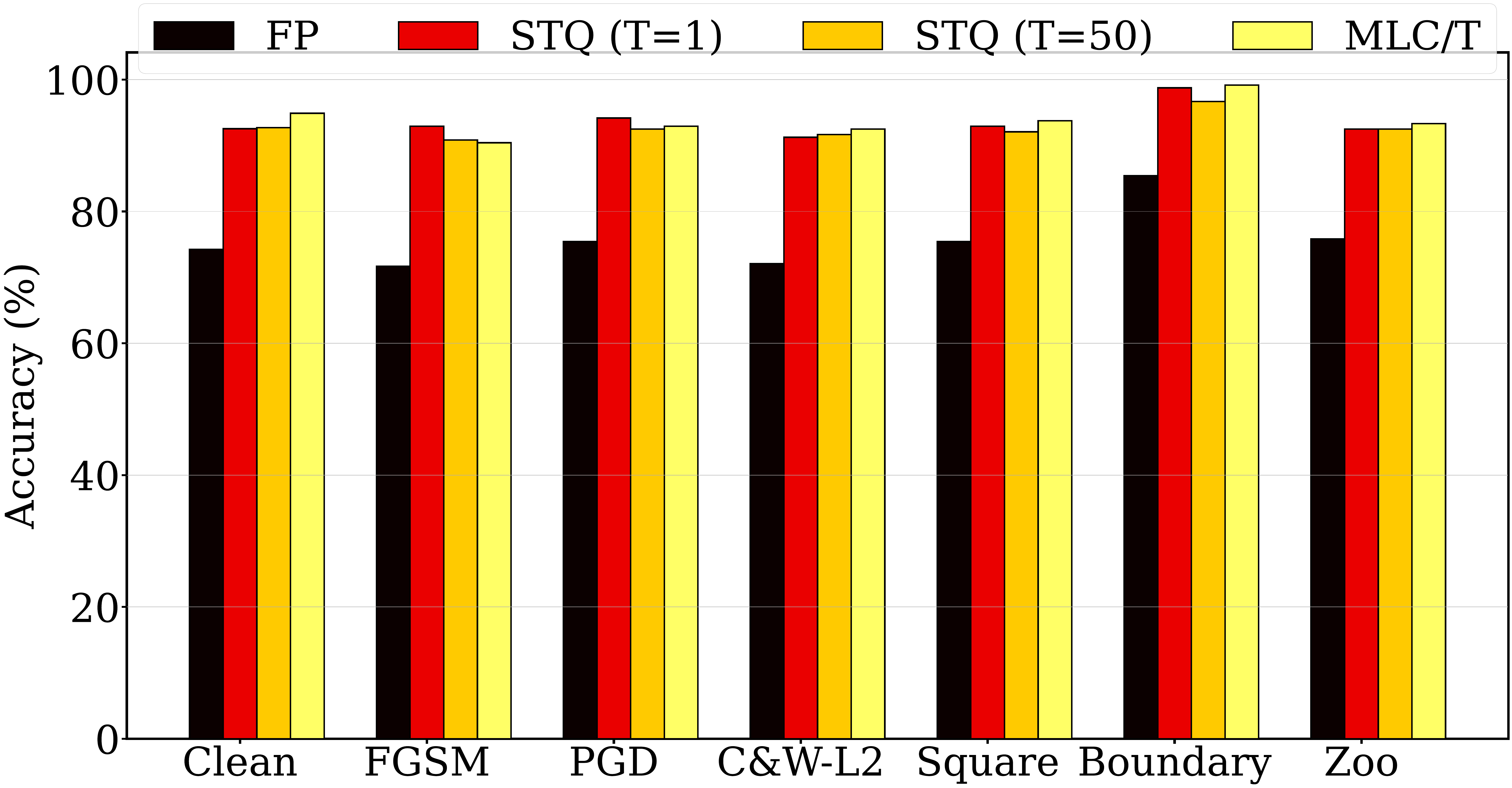}
    \caption{SVHN}
    \label{fig:svhn_kf}
  \end{subfigure}
  \begin{subfigure}[b]{0.70\linewidth}    
    \centering
   \includegraphics[width=0.99\linewidth]{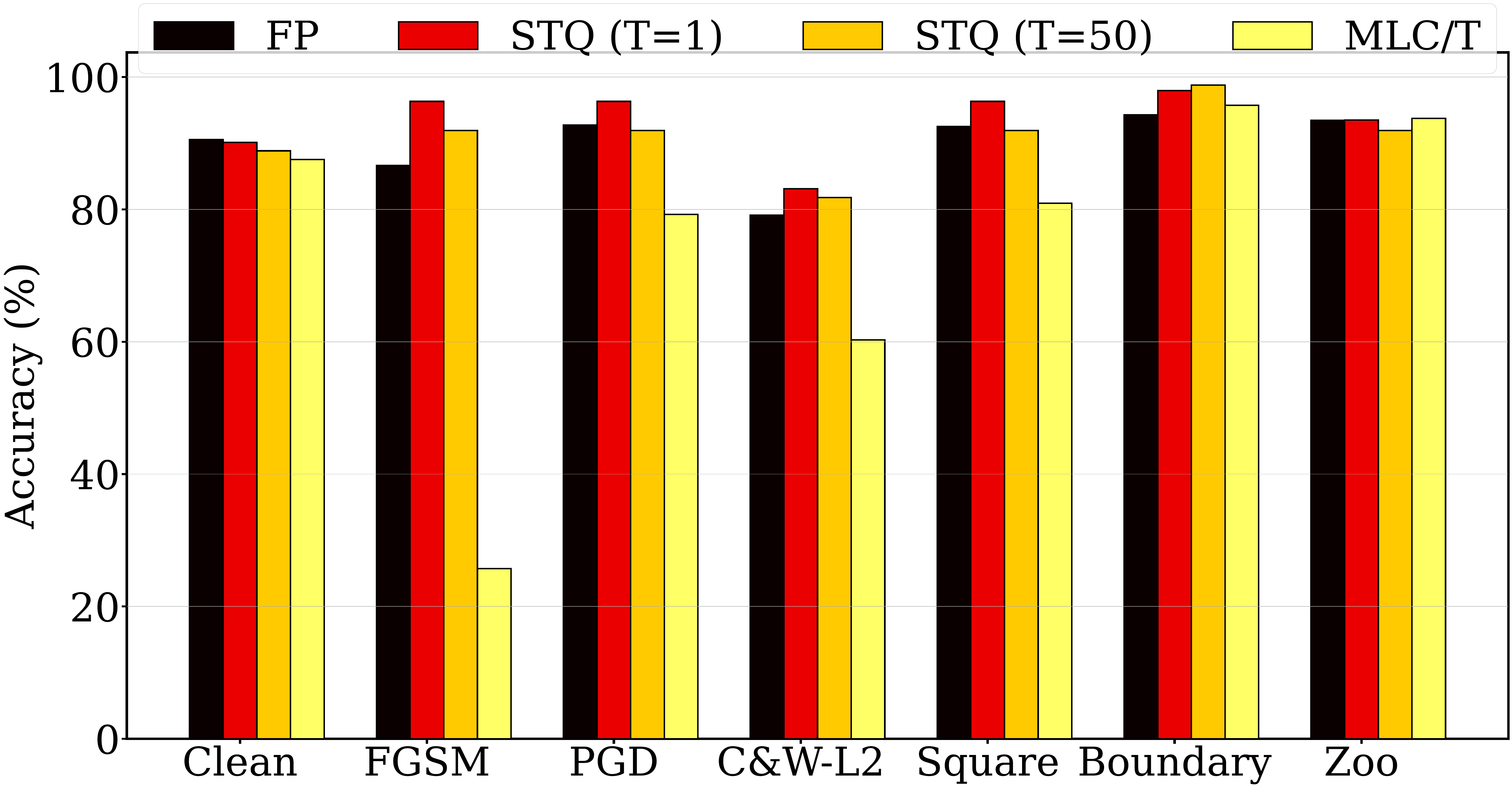}
    \caption{GSC}
    \label{fig:inf_c}
  \end{subfigure}
\caption{Models robustness (top-1 test accuracy) comparison for CIFAR-10, SVHN, and GSC datasets for a clean setting and across multiple white-box (FGSM, PGD, C\&W) and black-bow-attacks (Square, Boundary, ZOO). }
\label{fig:fig6}
\end{figure*}

Figure~\ref{fig:fig6} provides an analysis of the resilience of the proposed STQ model, as evaluated through top-1  average test accuracy for $5$ folds cross validation, under various adversarial attacks. These attacks encompass both white-box methodologies (FGSM, PGD, and C\&W-L2) and black-box techniques (Square, Boundary, and Zoo). 
To facilitate a comprehensive comparison, the achieved results include the performance of two MLC/T benchmark models: ResNetv1, specifically designed for image classification tasks, trained on CIFAR-10 and SVHN datasets, and DS-CNN, optimized for Keyword Spotting and trained on the GSC dataset, as documented in~\cite{MLperf}. 
The $K$-fold cross validation results are varying with datasets and types of attacks, although the overall average results demonstrate general consistency of STQ, with categorical cross entropy loss. Particularly for the GSC dataset, at which STQ tends to outperform the MLC/T for both clean and attack samples. These results demonstrate the performance capability of our STQ model, being a robust and effective model suitable to be deployed into resource-constrained edge devices.

As highlighted, the accuracy of the STQ model (basic implementation and including the defensive distillation technique, for a temperature equal to $50$), as well as the FP and two MLC/T benchmark models, are computed across multiple adversarial attacks. It can be noted that in the case of CIFAR-10, the ResNetV1 model exhibits in general superior performance in both clean data setting and adversarial attack setting.


Notably, STQ (without defensive distillation, $T = 1$) outperforms ResNetV1 in detecting FGSM, PGD, and Square attacks on the SVHN dataset but performs less effectively against the other adversarial attacks. This observation is intriguing, suggesting that the integration of STQ with JR can surpass existing models in countering specific attacks on certain datasets. 
This is further substantiated by comparing STQ with the GSC model, wherein the performance accuracy of STQ across all tested adversarial attacks are demonstrated. 
Particularly, STQ excels in FGSM, PGD, C\&W-L2 Square, and boundary attacks. This analysis demonstrates that varying the temperature parameter for STQ across different datasets can enhance its performance, particularly in the context of the GSC and SVHN datasets, where STQ's inherent JR mechanism proves to be advantageous.


\begin{figure*} 
\centering
  \begin{subfigure}[b]{0.45\linewidth}
    \centering
    \includegraphics[width=0.99\linewidth]{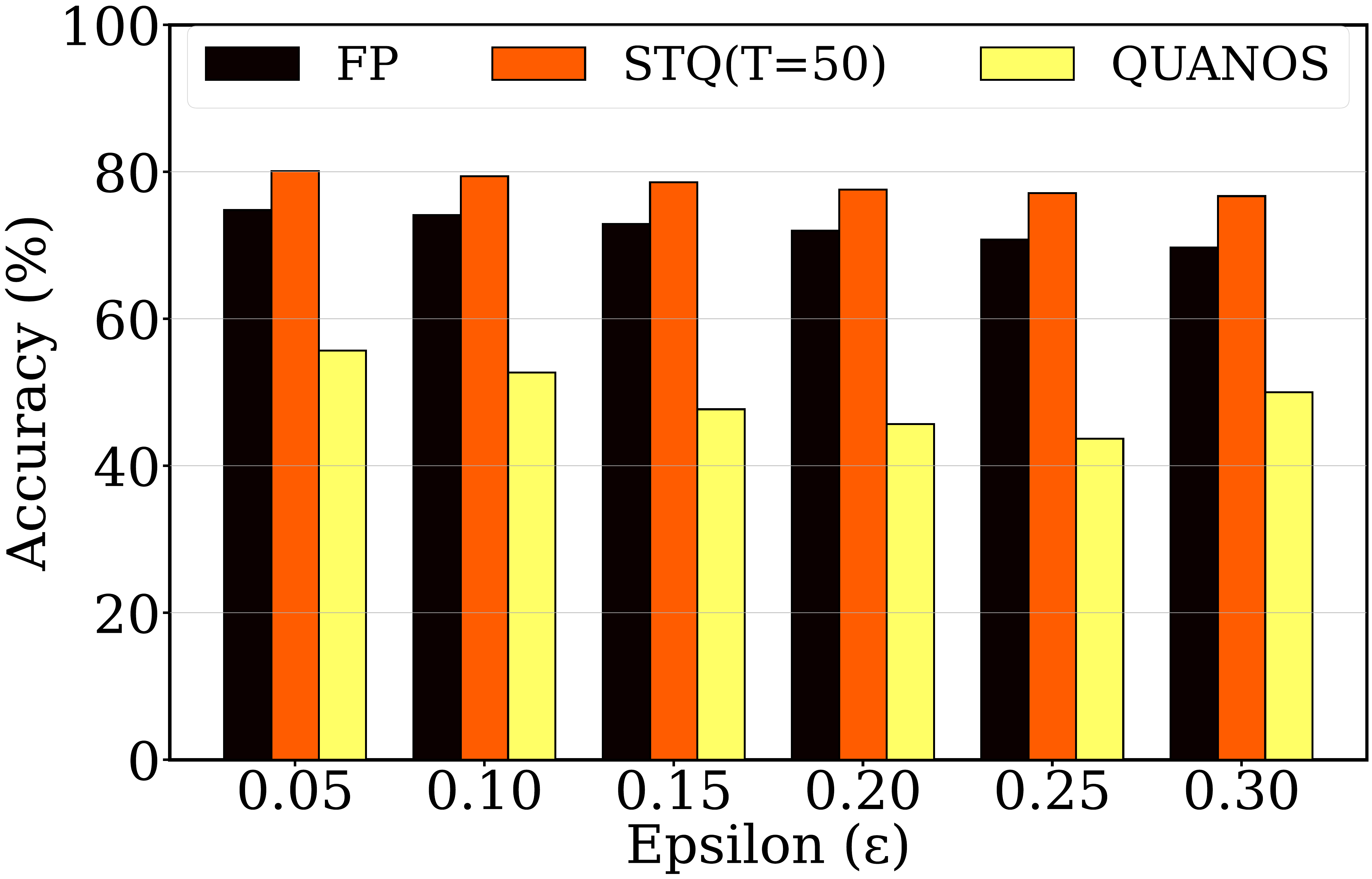}
    \caption{FGSM}
    \label{fig:inf_a}
  \end{subfigure}
  \begin{subfigure}[b]{0.45\linewidth}
    \centering
    \includegraphics[width=0.99\linewidth]{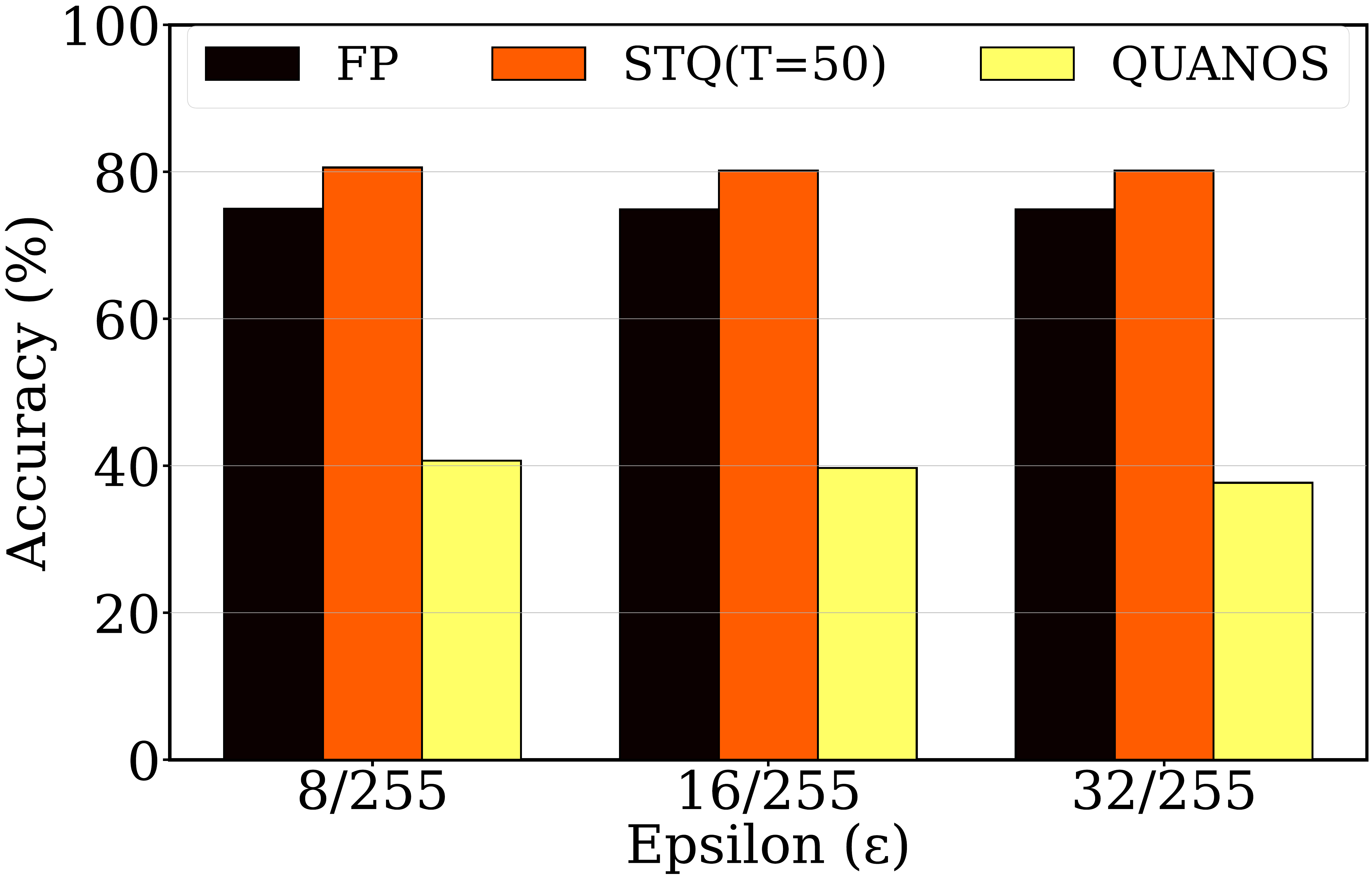}
    \caption{PGD}
    \label{fig:inf_b}
  \end{subfigure}
\caption{Robustness comparison of our FP and STQ models against Quanos~\cite{panda2020quanos} for various FGSM and PGD perturbation strengths.}
\label{fig:frameworks}
\end{figure*}

\begin{figure*} 
\centering
  \begin{subfigure}[b]{0.47\linewidth}
    \centering
    \includegraphics[width=0.99\linewidth]{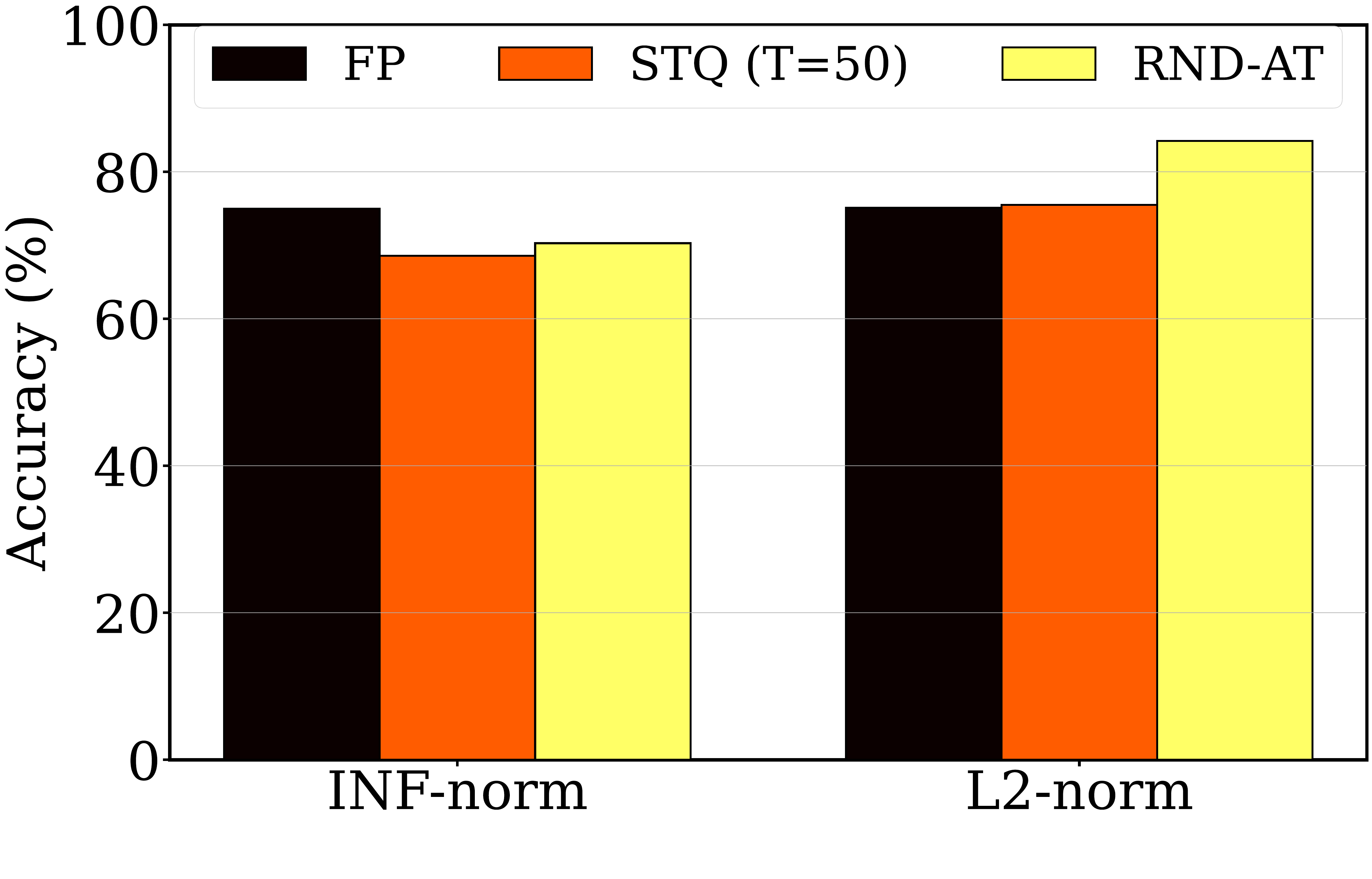}
    \caption{Square}
    \label{fig:inf_a}
  \end{subfigure}
  \begin{subfigure}[b]{0.48\linewidth}
    \centering
    \includegraphics[width=0.99\linewidth]{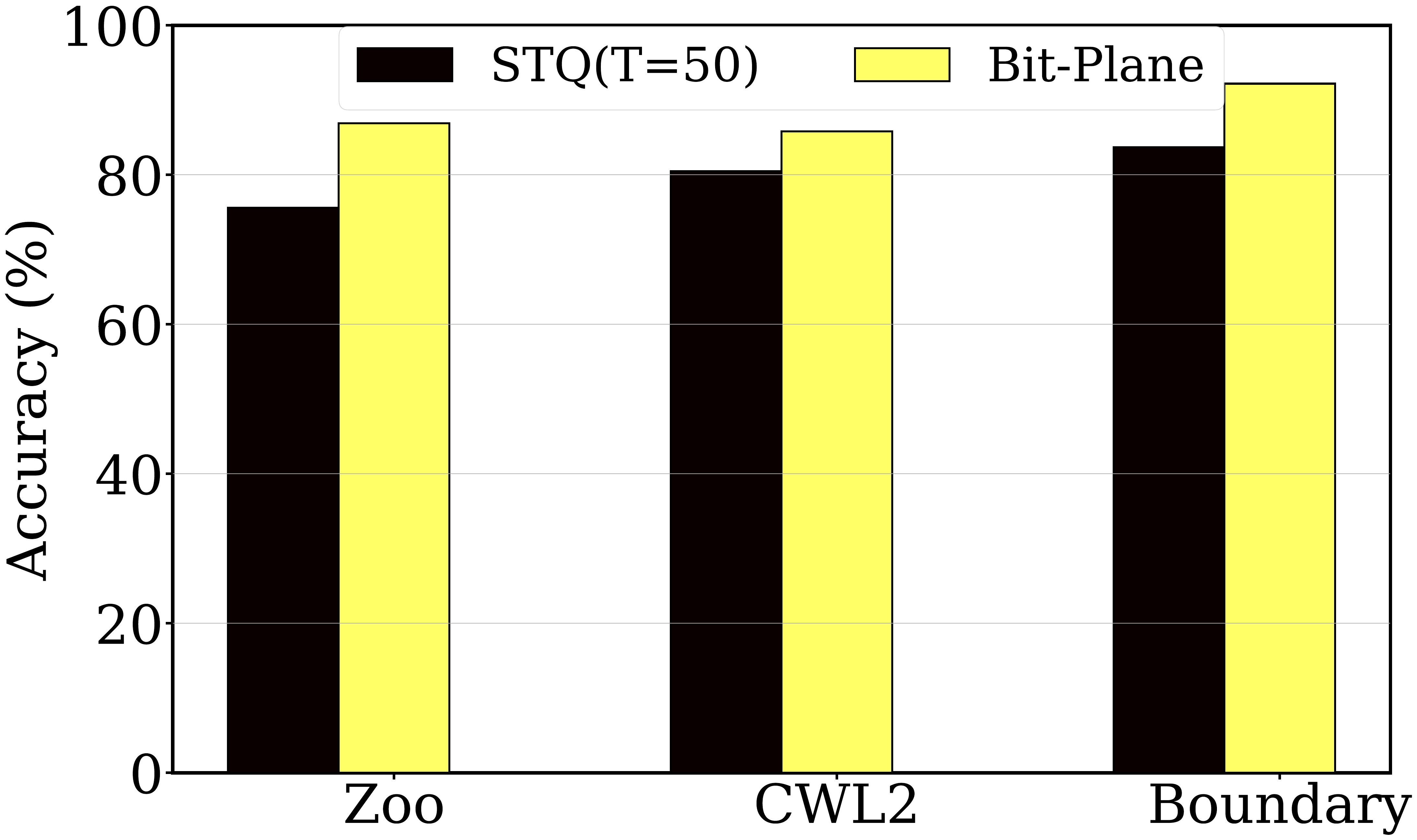}
    \caption{Square variants Attacks}
    \label{fig:inf_b}
  \end{subfigure}
\caption{Robustness comparison of our STQ models against RND-AT~\cite{qin2021random} for Square and other variation attacks.}
\label{fig:square}
\end{figure*}

Figure~\ref{fig:frameworks} offers a comparative analysis between the FP Keras version proposed and STQ-T50 models with QUANOS topology baseline model~\cite{panda2020quanos} across various FGSM and PGD perturbation strengths. 
Additionally, Figure~\ref{fig:square}(a) presents a comparison of FP, STQ-T50, and RND-AT~\cite{qin2021random} concerning different Square~\cite{square} variations, while Figure~\ref{fig:square}(b) focuses on C\&W-L2, Zoo and Boundary attacks. 
The STQ variant considered in these assessments is STQ-T=50, as it demonstrates superior accuracy with CIFAR-10 when compared to STQ-T=1. These evaluations were conducted using $10,000$ testing samples from the CIFAR-10 dataset, except for the C\&W-L2, Zoo and Boundary attacks were $2000$ random samples were utilized~\cite{liu2022defending}. 
The results indicate that both FP and STQ models clearly surpass QUANOS in detecting FGSM and PGD attacks across various strengths. However, they exhibit slightly lower performance than RND-AT when dealing with Square attacks.

Finally, a comparative analysis was carried out involving MobileNetV2 in conjunction with the proposed FP and STQ models. This comparison encompassed aspects such as accuracy and memory footprint requirements. It is worth noting that the optimizer employed during MobileNetV2 training was Stochastic Gradient Descent (SGD), as recommended for enhancing model convergence~\cite{song2021communication}.

Table~\ref{table:stqevalcomp} shows a performance comparison between our model (FP and STQ-T50) and MobileNetV2 (FP and to which the STQ-T50 methodology has been applied) with regard to their top-1 test accuracy when tested on clean inputs from the CIFAR-10 dataset. 
The table also includes their FLASH MCU memory requirements. 
The results reveal that MobileNetV2-STQ-T50 demonstrate reduction in accuracy compared to the MobileNetV2. This indicates that the application of STQ methodoloy in MobileNetV2 leads to a reduction in accuracy. However, this reduction in accuracy is accompanied by a significant decrease in flash memory requirement, when compared to their MobileNetV2 counterparts. Indeed the MobileNetV2 is very far to be deployable on a 2MB FLASH MCU such as the high end STM32H7. 
Therefore it was worth applying the STQ methodology to MobileNet-V2 to save $308$ KB of the FLASH memory required to implement the model on an edge device and, at the same time, increase the accuracy by 4.66\% compared to the proposed multimodal model. A task that remains open at this point in time is the quantitative resilience of MobileNetV2-STQ-T50 to adversarial attacks.

\begin{table}[!t]
\caption{Performance (top-1 accuracy) comparison with benchmarks on CIFAR-10 clean datasets.}
\begin{center}
\label{table:stqevalcomp}
\begin{tabular}{|c|c|c|c|c|c|c|}
\hline
\textbf{\multirow{2}{*}{Model}} & \textbf{Flash} & \textbf{Accuracy}  \\
& \textbf{(KB)} & \textbf{(\%)}   \\ \hline
{FP} & {4496} & {75.20}   \\ \hline
{STQ-T50} & {410} & {80.57}   \\ \hline
{MobileNetV2} & {8796} & {88.58}  \\ \hline
{MobileNetV2-STQ-T50} & {102} & {85.23}  \\ \hline
\hline
\end{tabular}
\end{center}
\end{table}

\section{Conclusion} 
\label{conc}

This paper investigated the robustness of a Deeply Quantized Machine Learning model against various white-box and black-box adversarial attacks. The deep quantization feature of QKeras was considered to create a memory optimized, accurate and adversarially robust quantized model. 
This paper proved that the reason of increased robustness in deeply quantize ML is its inherent similarity with a defense technique, Jacobian Regularization (JR). 
As demonstrated, the proposed Stochastic Ternary Quantized (STQ) model, with quantization-aware training procedure introducing per-layer JR, was more robust than industry adopted MLCommons/Tiny benchmarks when facing several adversarial attacks. 
In fact, the stochastic quantization scheme was effective for model compression with support for robustness against adversarial attacks. This robustness was experimentally proven by observing its accuracy under various adversarial attacks utilizing two image (CIFAR-10 and SVHN) datasets and one audio (GSC) dataset. 
This is relevant in the context of deploying efficient and effective DNN models in resource-constrained environments, with limited capabilities. These initial results suggest further exploration of other sophisticated white-box and black-box attacks with different attack strengths. 
Future work will further assess the effectiveness of the proposed STQ QKeras model against new and latest attacks and applied to state-of-the-art models. 



\section*{Acknowledgment}

This work was supported by the PETRAS National Centre of Excellence for IoT Systems Cybersecurity, funded by the UK EPSRC under grant number EP/S035362/1.


\bibliographystyle{ACM-Reference-Format}
\bibliography{bibfile}

\end{document}